# Multimodal Fusion: A Review of Breast Cancer Advanced and Traditional Modalities

Jawaher Al-Tuweity* and João Paulo Papa

Computing Department, Faculty of Sciences, University of São Paulo State
UNESP, Bauru, 17033-360, São Paulo, Brazil.

*Corresponding author(s). E-mail(s): jawaher.altuweity@unesp.br;

**Abstract**

**Purpose:** Multimodal medical imaging fusion (MMIF) is an increasingly important approach to breast cancer detection (BCD), as different imaging modalities rely on different physical principles and provide complementary information. No single modality can capture all the structural, functional, and biological information needed for accurate diagnosis and characterization. This review examines the main imaging modalities used in breast cancer detection and the role of multimodal fusion in improving diagnostic performance.

**Methods:** This review examines more than 15 imaging modalities used in breast cancer detection, including conventional techniques such as mammography, ultrasound, and magnetic resonance imaging (MRI), as well as emerging approaches such as thermal imaging, photoacoustic imaging, microwave imaging, and multimodal microscopy. It discusses their underlying principles, complementary contributions, and applications in multimodal fusion, with particular attention to recent advances in artificial intelligence (AI), deep learning, and transformer-based models.

**Results:** Mammography, ultrasound, and MRI remain among the most widely used modalities, providing complementary structural and functional information. Recent AI advances have enabled fusion at the data, feature, and decision levels, with applications in detection, classification, segmentation, prognosis, and personalized cancer management. Multimodal approaches increasingly combine imaging data with clinical, histopathological, and genomic information. However, important challenges remain, including limited availability of large and diverse multimodal datasets, variations in imaging protocols, missing modalities, limited external validation, and the need for interpretable and clinically reliable AI models.

**Conclusion:** MMIF offers promising opportunities to improve breast cancer detection and characterization by combining complementary imaging and non-imaging information. Future research should prioritize robust fusion strategies, standardized and diverse datasets, multicenter validation, explainable and privacy-preserving AI, and the integration of imaging with clinical and molecular data to facilitate reliable translation into clinical practice.

# 1 Introduction

Precision and timeliness in breast cancer detection are important for improving patient outcomes. Traditional diagnostic methods have predominantly relied on unimodal approaches, but recent advancements in medical data analytics have enabled the integration of diverse data sources beyond conventional imaging techniques Abdullakutty et al. (2024). Multimodal medical imaging fusion has become a cornerstone in advancing breast cancer detection, offering improved diagnostic accuracy by integrating complementary information from various imaging modalities such as mammography, ultrasound, MRI, PET, and histopathology. Recent literature reviews and surveys highlight the transformative potential of combining these modalities with advanced computational methods—including deep learning, machine learning, and explainable AI to enhance early detection, classification, and personalized treatment planning for breast cancer Abdullakutty et al. (2024); Guo et al. (2017); Cerdas et al. (2025); Figueiras et al. (2025); H et al. (2025); Houssein et al. (2021); Hsu et al. (2025); Jha et al. (2025); Lima et al. (2019); Ming et al. (2020); Rahman et al. (2025); Rezaei (2021); Sushanki et al. (2023). Foundational works categorize fusion strategies into early, intermediate, and late fusion approaches, each with unique strengths in handling heterogeneous data and improving robustness against noisy or incomplete inputs Abdullakutty et al. (2024); Mashekova et al. (2025). The integration of non-image data (e.g., genomics, clinical records) with imaging is an emerging trend that promises further gains in diagnostic precision but remains underexplored Abdullakutty et al. (2024); Hou et al. (2025). Despite significant progress especially with AI-assisted multimodal frameworks challenges persist regarding data standardization, model interpretability, clinical validation, and the need for large-scale public datasets Figueiras et al. (2025); Hou et al. (2025); Jha et al. (2025); Rezaei (2021).

A 67% in recent reviews and clinical studies, some in high-impact journals, consistently report that multimodal medical imaging fusion improves breast cancer detection accuracy with multimodal imaging and AI integration. While 33% narrative and technical reviews highlight promising results but emphasize the need for large-scale clinical trials and standardization before routine adoption Alzamil et al. (2025); Hsu et al. (2025); Maruf et al. (2025); Zheng et al. (2023).

This review synthesizes the current state of multimodal imaging fusion for breast cancer detection traditional and advance modalities by examining recent surveys, clinical applications, challenges and future research directions.

## 1.1 Breast Cancer Detection Modalities

Breast cancer is the most frequently diagnosed cancer and the leading cause of cancerrelated mortality among women worldwide, accounting for more than 2.3 million new cases and over 680,000 deaths annually. Breast cancer is the most common cancer, which resulted in the death of 700,000 people around the world in 2020 Wilkinson and Gathani (2022).Various imaging modalities have been utilized to detect and analyze breast cancer. However, the manual detection of cancer from large-size images produced by these imaging modalities is usually time-consuming and can be inaccurate. Early and accurate detection of breast cancer plays a important role in improving the prognosis bringing the patient survival rate to 50%. Recently, some artificialintelligence-based approaches such as deep learning algorithms have shown remarkable advancements in early breast cancer diagnosis Madani et al. (2022). Early detection and accurate characterization of breast lesions is important for improving survival outcomes, enabling personalized treatment planning, and reducing unnecessary biopsies and overtreatment. Medical imaging is a central role in breast cancer screening, diagnosis, staging, and treatment monitoring, with modalities such as mammography, ultrasound, magnetic resonance imaging (MRI), and positron emission tomography (PET) being routinely used in clinical practice Lehman et al. (2015); Mann et al. (2019); Sung et al. (2021).

Despite significant advances in imaging technology, no single modality can comprehensively capture the complex morphological, functional, and molecular characteristics of breast tumors. Mammography provides high spatial resolution and is widely used for population screening but has reduced sensitivity in dense breast tissue. Ultrasound offers real-time imaging and is useful for lesion characterization but is operatordependent. MRI provides high sensitivity and multiparametric tissue characterization, while PET offers functional metabolic information at the whole-body level. These complementary modalities highlight the need for integrative frameworks capable of leveraging heterogeneous imaging and non-imaging data sources Fuller et al. (2015); Mann et al. (2019); Qi et al. (2024).

Breast cancer detection has multiball imaging modalities, see Figure 1 and Table 1, clinically has different Insights, Mammography remains the gold standard for population screening. MRI is the most sensitive modality for breast cancer detection and therapy monitoring. Ultrasound is essential for dense breasts and biopsy guidance. PET-based modalities provide metabolic and systemic disease information. Multimodal fusion (MRI + MM + US + Clinical) provides the highest diagnostic accuracy in AI systems. Histopathology remains the clinical ground truth for diagnosis. Radiogenomics is emerging for personalized treatment prediction. Table 2 show the BCD modalities in summery and role comparison.

The evolution of breast cancer detection modalities has progressed from conventional imaging techniques such as mammography and ultrasound to advanced modalities including MRI and PET/CT. Recent advances in artificial intelligence have enabled the integration of multimodal data, combining imaging, clinical, and genomic information to support personalized diagnosis and treatment planning.

Given the continuous and rapid evolution of medical imaging technologies, the categorization of imaging modalities in this work into “traditional” and “advanced” should be interpreted primarily as a reflection of their current clinical adoption and common usage rather than as a strict technological boundary. Such classification does not fully capture the ongoing developments within individual modalities, where technological innovations, novel acquisition protocols, and enhanced imaging techniques may substantially transform the capabilities of a conventionally established modality into a more advanced diagnostic approach.

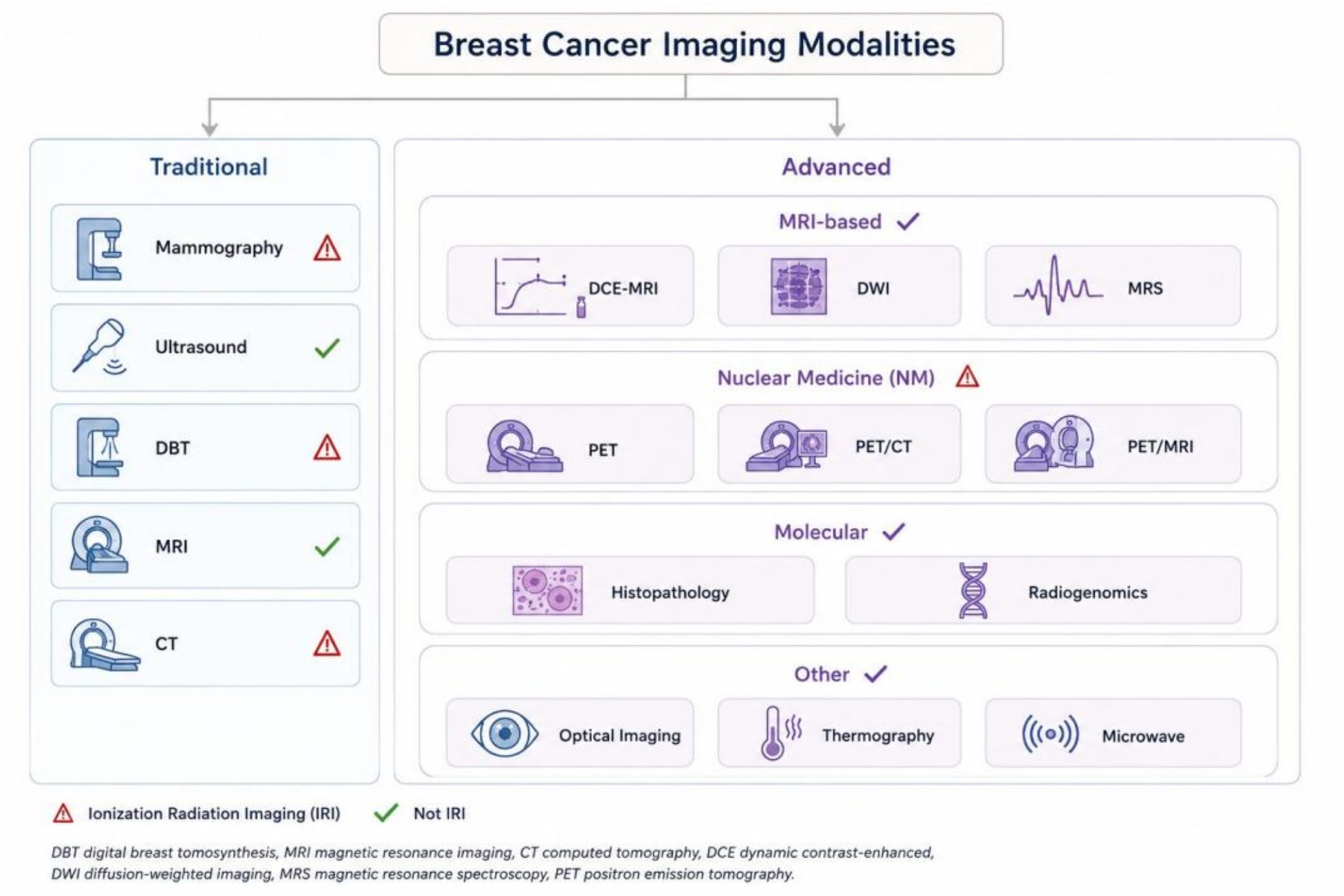

**Fig. 1** Illustration for BCD Traditional and Advanced modalities

## 1.2 Multimodel Medical Imaging Fusion MMIF and Challenges

In recent years, artificial intelligence (AI) and deep learning have revolutionized medical image analysis, enabling automated detection, segmentation, and classification of breast lesions with performance comparable to or exceeding human experts. Deep convolutional neural networks (CNNs) and transformer-based models have demonstrated strong performance in mammographic screening, MRI-based tumor segmentation, and ultrasound lesion classification Geras et al. (2018); McKinney et al. (2020); Shen et al. (2019); Yala et al. (2019). However, most early AI systems were developed for single-modality imaging, limiting their ability to exploit complementary cross-modal information.

Data fusion, referred to as multimodal integration, aims to combine information from multiple imaging modalities and heterogeneous data sources to improve diagnostic accuracy, robustness, and clinical interpretability. Fusion strategies Figure 2 are commonly categorized into image-level (pixel-level), feature-level, and decision-level fusion, with recent deep learning approaches enabling intermediate and hierarchical fusion across multiple network layers Baltruschat et al. (2019); Ragab et al. (2019); Suzuki (2017).

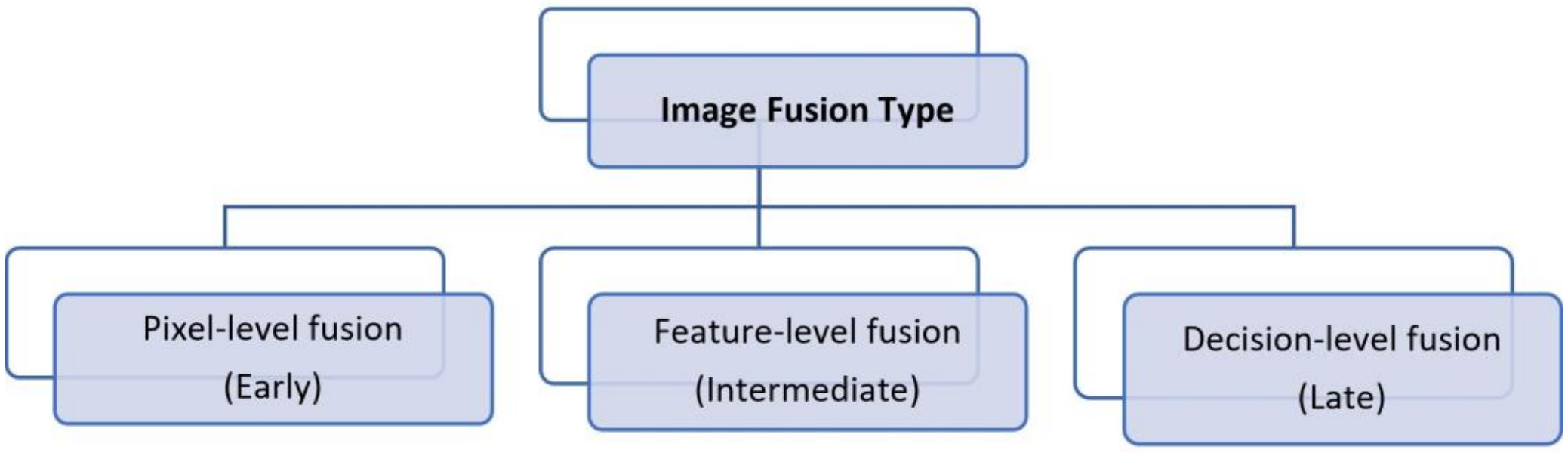

**Fig. 2** Image Fusion Type

Multimodal fusion has shown promising results in breast cancer detection, risk prediction, and therapy response assessment by integrating mammography, ultrasound, MRI, clinical variables, and genomic biomarkers Agarwal (2021); Li et al. (2024); Wang et al. (2023); York et al. (2026).

Beyond imaging modalities, emerging radiogenomic and multimodal AI frameworks integrate imaging phenotypes with molecular and clinical data to enable precision oncology. Large-scale multimodal datasets such as The Cancer Genome Atlas (TCGA) have facilitated the development of predictive models for molecular subtyping, prognosis, and treatment response. Moreover, foundation models, self-supervised learning, and transformer-based architectures have recently been introduced to unify multimodal representation learning and improve generalization across institutions and imaging protocols Liang (2024); Qu et al. (2024); Zhang et al. (2026).

Despite these advances, several challenges remain in multimodal breast cancer imaging fusion, including limited availability of large annotated multimodal datasets, domain shift across institutions and scanners, lack of standardized benchmarks, interpretability concerns, and barriers to clinical translation. Existing surveys have primarily focused on general multimodal fusion or single-modality breast imaging AI, with limited emphasis on systematic taxonomy, clinical relevance, and comparative analysis of fusion strategies specific to breast cancer imaging Adam et al. (2023); Litjens et al. (2017).

## 1.3 Classical Fusion in Medical Imaging

Classical data fusion approaches in medical imaging include image-level, featurelevel, and decision-level fusion. Image-level fusion methods combine multiple imaging sequences or modalities through registration and pixel-wise integration, often using wavelet transforms or statistical fusion frameworks. Feature-level fusion concatenates handcrafted or learned features extracted from different modalities, followed by machine learning classifiers such as support vector machines or random forests.

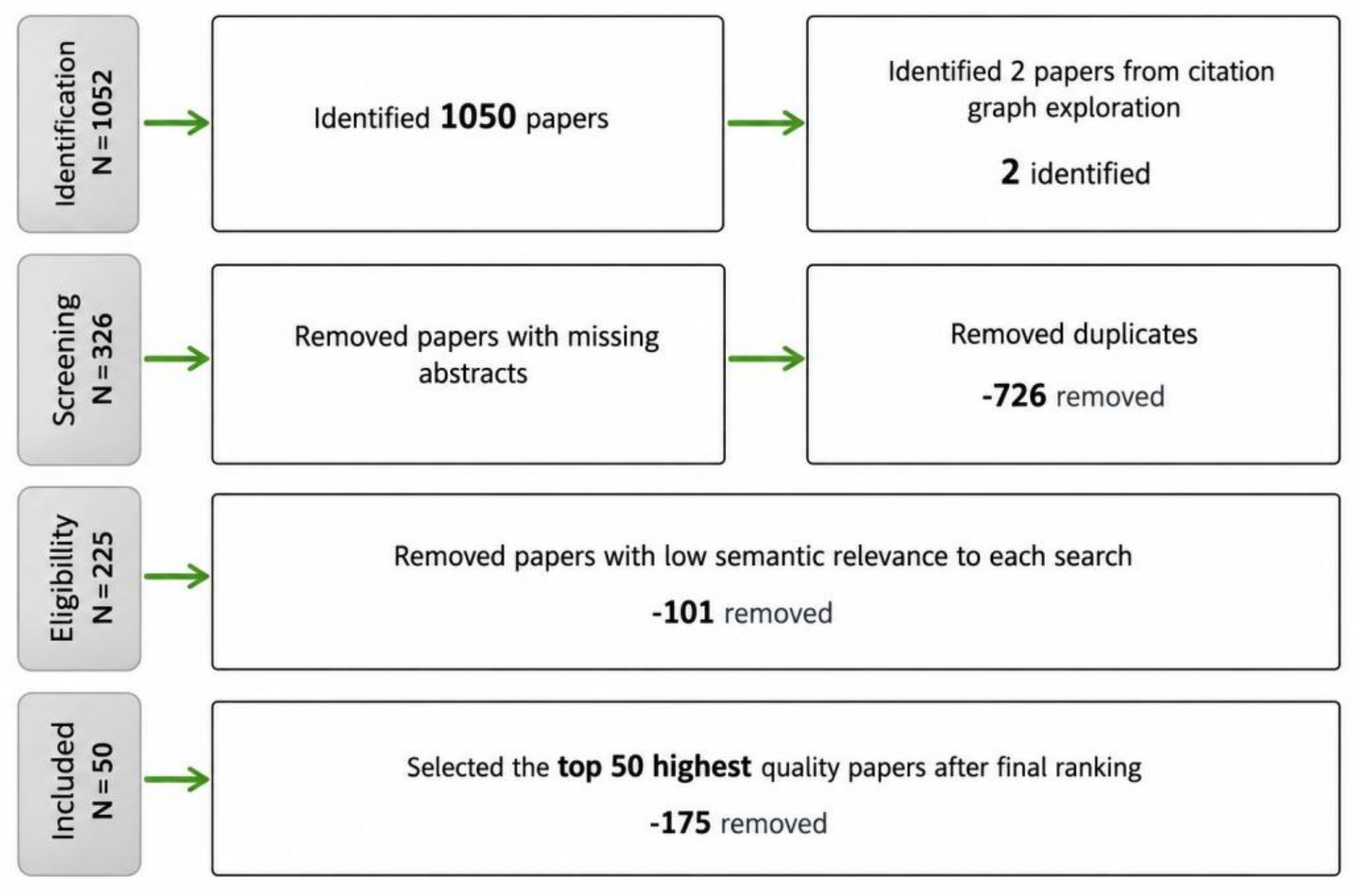

**Fig. 3** PRISMA-style literature screening flowchart

Decision-level fusion integrates outputs from independent models using ensemble strategies or Bayesian frameworks Obayya et al. (2022) Knights et al. (2026).

Early multimodal breast cancer studies combined mammography and ultrasound or MRI using handcrafted radiomic features. Radiomics-based fusion frameworks demonstrated improved diagnostic performance compared to single-modality radiomics, but were limited by handcrafted feature instability and domain dependence.

## 2 Methodology

A comprehensive search was conducted across over 170 million research papers in Consensus database including Semantic Scholar, PubMed, in addition to IEEE, Nature, Elsevier journals, RSNA and other databases using targeted queries related to multimodal medical imaging fusion for breast cancer detection cover the period between 2017-2025.

Figure 3 show the PRISMA-style literature screening flowchart showing the identification, screening, eligibility assessment, and final inclusion of studies. A total of 1,050 papers were identified through the literature search, with two additional papers identified through citation graph exploration. After removal of duplicate records, papers with missing abstracts, and studies with low semantic relevance, the final ranking resulted in the inclusion of the 50 highest-quality papers.

**Table 1** BCD Medical Imaging Comparison of Advanced and Traditional Modalities

| Modality | Abbreviation | Physical Principle | Primary Clinical Use in Breast Cancer | Strengths | Limitations | Typical Clinical Tasks | Spatial Resolution |
|---|---|---|---|---|---|---|---|
| **Mammography** | MM or MG | X-ray attenuation imaging | Primary screening | Low cost, high spatial resolution | Low sensitivity in dense breasts | Screening, lesion detection | High (50–100 μm) |
| **Digital Breast Tomosynthesis** | DBT | 3D X-ray projections | Advanced screening | Reduces tissue overlap | Higher dose than MG | Screening, lesion localization | High |
| **Ultrasound** | US | High-frequency acoustic waves | Adjunct to mammography | No radiation, differentiates cystic from solid lesions | Operator-dependent | Diagnosis, lesion characterization | Moderate |
| **Contrast-Enhanced Ultrasound** | CEUS | Microbubble contrast | Tumor vascularity assessment | Real-time perfusion imaging | Limited penetration, contrast safety considerations | Tumor characterization, therapy monitoring | Moderate |
| **Magnetic Resonance Imaging** | MRI | Magnetic resonance | High-risk screening, staging | High sensitivity, multiparametric imaging | High cost, long acquisition time | Detection, staging, therapy response | High (~1 mm) |
| **Dynamic Contrast-Enhanced MRI** | DCE-MRI | Gadolinium perfusion imaging | Tumor vascularity assessment | High sensitivity | Contrast risk, cost | Diagnosis, pCR prediction | High |
| **Diffusion-Weighted Imaging** | DWI-MRI | Water diffusion measurement | Tumor cellularity assessment | No contrast needed | Susceptibility artifacts | Tumor characterization, therapy response | Moderate |
| **Magnetic Resonance Spectroscopy** | MRS-MRI | Metabolic spectroscopy | Tumor metabolic profiling | Metabolic information | Low spatial resolution | Metabolic characterization, exploratory biomarker assessment | Low |
| **Computed Tomography** | CT | X-ray computed tomography | Metastasis detection | Whole-body imaging | Poor soft-tissue contrast | Staging | Moderate |
| **Positron Emission Tomography** | PET | Radiotracer metabolic imaging | Staging, metastasis detection | Functional imaging | High cost, radiation | Metastasis detection, therapy monitoring | Low |
| **PET/CT** | PET/CT | PET + CT fusion | Whole-body staging | Anatomical and functional fusion | High radiation dose | Metastasis detection, treatment monitoring | Moderate |
| **PET/MRI** | PET/MRI | PET + MRI fusion | Advanced research, staging | High soft-tissue contrast and metabolic information | Very high cost | Precision oncology, staging | High |
| **Optical Imaging** | OI | Near-infrared light | Research vascular imaging | Low cost, real-time imaging | Low penetration depth | Research angiogenesis assessment | Low |
| **Thermography** | TG | Infrared thermal imaging | Investigational adjunct | Non-invasive, no radiation | Low specificity, surface temperature information | Research applications | Low |
| **Histopathology** | H&E | Microscopic examination of biopsy tissue | Gold-standard tissue diagnosis | Cellular-level detail | Invasive biopsy | Cancer grading, subtype classification | Very high |
| **Radiogenomics** | RG | Integration of imaging and genomic data | Precision oncology research | Potential to predict molecular characteristics | Data scarcity, analytical complexity | Subtype prediction, therapy response | Not applicable |
| **Microwave Imaging** | MWI/MBI | Dielectric property contrast | Experimental breast tumor detection | Non-ionizing, potentially low-cost and portable | Limited spatial resolution, clinical validation | Experimental tumor detection | Experimental; system-dependent |

# 3 Literature Review LR

Recent literature demonstrates a growing interest in multimodal medical imaging fusion (MMIF) for breast cancer detection (BCD), diagnosis, prognosis, and treatment monitoring. The LR selection papers have investigated the traditional and advanced imaging modalities, including mammography, digital breast tomosynthesis (DBT), ultrasound (US), magnetic resonance imaging (MRI), positron emission tomography (PET), computed tomography (CT), histopathology, thermography, and emerging techniques. The literature increasingly emphasizes the integration of complementary imaging information with genomic, clinical, and pathological data to improve diagnostic performance and support precision oncology. Fusion approaches can be broadly categorized as early/input, intermediate/feature, and late/decision-level fusion, with deep learning architectures incorporating attention mechanisms, encoder–decoder networks, graph neural networks, and transformers Abdullakutty et al. (2024); Li et al. (2024); Huang et al. (2020); Yadav and Yadav (2020); Khan et al. (2023).

Mammography, DBT, US, and MRI remain the principal modalities for breast cancer screening and diagnosis. US is particularly valuable because of its accessibility, absence of ionizing radiation, and ability to characterize lesions, although its performance is operator-dependent Guo et al. (2017). DBT provides three-dimensional information and reduces tissue overlap, while deep learning has increasingly been investigated for automated DBT interpretation and screening Bai et al. (2021). MRI offers high sensitivity and multiparametric information through techniques such as dynamic contrast-enhanced MRI (DCE-MRI) and diffusion-weighted imaging (DWI), supporting lesion detection, staging, and treatment-response assessment. Abbreviated MRI protocols have also been investigated as lower-cost and faster alternatives while maintaining high diagnostic performance Ewaidat and Ayasrah (2022); Mytych et al. (2025); Jannatdoust et al. (2025). PET/CT and PET/MRI provide complementary metabolic and anatomical information and are particularly useful for staging and metastatic disease assessment Ming et al. (2020); Pesapane et al. (2020).

Artificial intelligence (AI), machine learning (ML), and deep learning (DL) have become central to automated breast image analysis. CNNs, pretrained networks, transformers, and other DL architectures have been applied to detection, segmentation, classification, risk prediction, and treatment-response assessment across multiple modalities Houssein et al. (2021); Abhisheka et al. (2023); Jiang et al. (2023); Luo et al. (2023); Radak et al. (2023). Recent studies indicate that multimodal approaches can provide more comprehensive information than single-modality systems, particularly when imaging is combined with clinical, genomic, or pathological data Li et al. (2024); Abdikenov et al. (2025). Self-supervised learning (SSL) has emerged as a promising strategy for reducing dependence on large labeled datasets by exploiting unlabeled mammography, US, MRI, DBT, and histopathology data, although applications to PET and multimodal fusion remain comparatively limited Figueiras et al. (2025). Federated learning and explainable AI (XAI) are also being explored to address privacy, interpretability, and clinical trust. Methods such as Grad-CAM, SHAP, and LIME provide additional insight into model predictions H et al. (2025); Abdullakutty et al. (2024).

Radiomics further extends multimodal analysis by extracting quantitative features that may support lesion characterization, molecular subtype assessment, lymphnode metastasis prediction, prognosis, and treatment-response evaluation Conti et al. (2020); Zhao et al. (2025).

However, the literature consistently identifies limitations related to small or imbalanced datasets, heterogeneous imaging protocols, lack of external validation, domain shift, incomplete multimodal data, limited interpretability, and insufficient standardization Sushanki et al. (2023); Sharafaddini et al. (2024); Ali et al. (2025); Mashekova et al. (2025). These issues and the challanges restrict generalizability and remain major barriers to clinical translation.

Alongside established modalities, emerging techniques such as contrast-enhanced mammography (CEM), microwave breast imaging, optical imaging, photoacoustic imaging, Raman spectroscopy, hyperspectral imaging, and molecular breast imaging are receiving increasing attention Coffey and Jochelson (2022); Bhargava et al. (2024); Hsu et al. (2025); Cerdas et al. (2025). Although these approaches offer potential advantages in tissue characterization, non-ionizing imaging, or functional assessment, their clinical adoption requires further technical development, standardization, multicenter validation, and cost reduction.

Overall, the reviewed literature shows a clear transition from unimodal image analysis toward multimodal, AI-driven breast cancer assessment. Bibliometric and review studies demonstrate rapidly increasing research activity, particularly since 2021, with major emphasis on deep learning, multimodal fusion, radiomics, explainability, and personalized medicine Hou et al. (2025); Jha et al. (2025). Nevertheless, a significant gap remains between promising experimental performance and clinically generalizable systems. see Table 3 Key claims and supporting evidence identified in LR papers.

**Table 2** Summary & Comparison of BCD Modalities Roles

| **Modality Category** | **Screening** | **Diagnosis** | **Staging** | **Therapy Monitoring** | **Prognosis Prediction** |
|---|---|---|---|---|---|
| Mammography / DBT | Yes | Yes | Limited | No | Emerging |
| Ultrasound | No | Yes | Limited | Yes | Emerging |
| MRI (DCE/DWI) | High-risk screening | Yes | Yes | Yes | Yes |
| PET / PET-CT / PET-MRI | No | Limited | Yes | Yes | Emerging |
| CT | No | Limited | Yes | Yes | No |
| Optical / Thermography | Research | Research | No | Research | No |
| Histopathology | No | Gold Standard | Yes | Yes | Yes |
| RadioGenomics | No | Research | No | Research | High Potential |

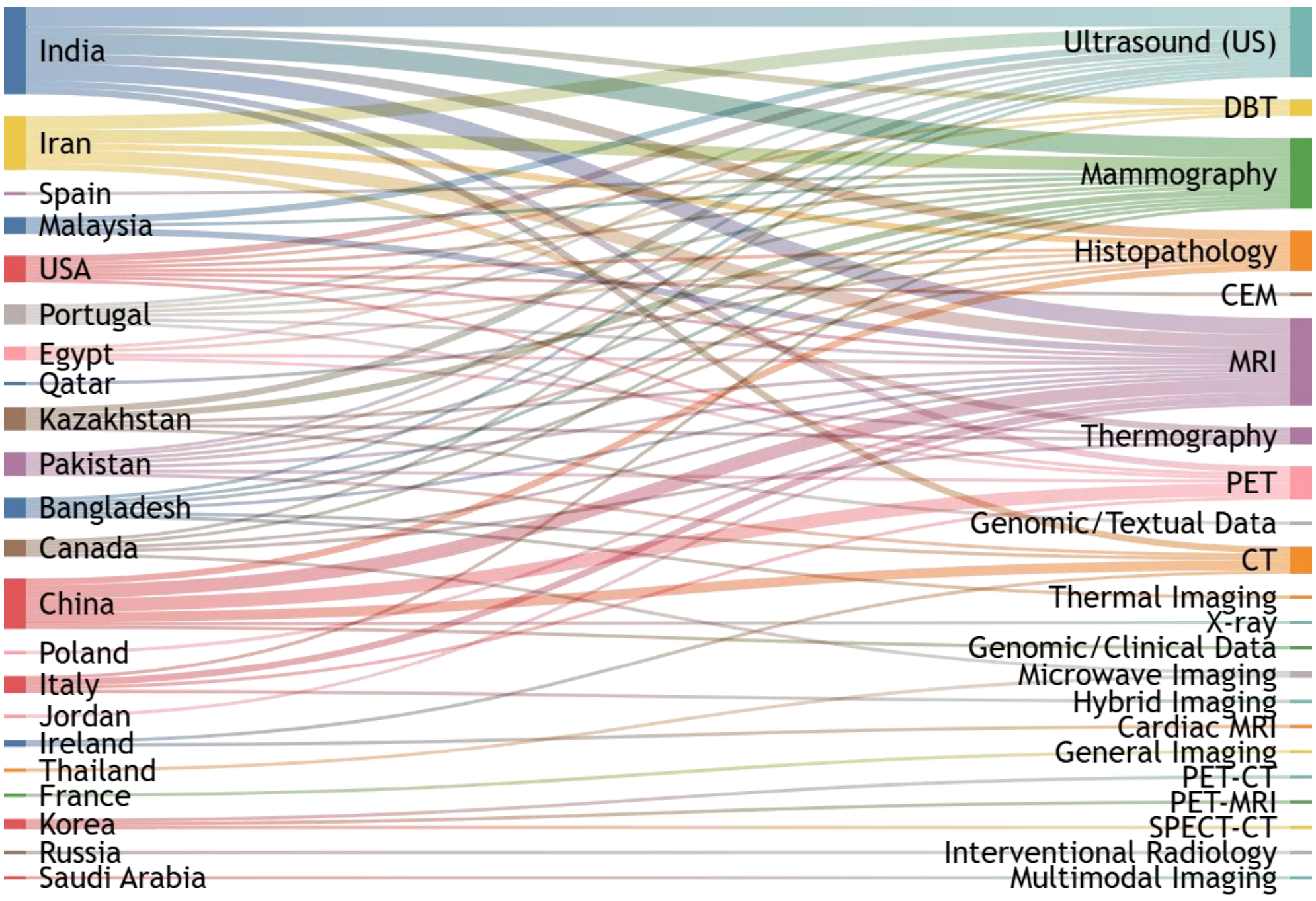

**Fig. 4** The Sankey diagram illustrates the distribution of breast cancer imaging modalities across different countries and modality combinations. Countries such as India, China, Iran, and the USA contribute the largest number of studies. Mammography emerges as the most widely used imaging modality, followed by magnetic resonance imaging (MRI) and ultrasound (US). The diagram also highlights emerging trends toward multimodal data fusion by integrating imaging data with genomic and clinical information, as well as the increasing adoption of advanced imaging techniques such as microwave imaging and hybrid imaging.

# 4 Results and Discussion

The literature consistently demonstrates that multimodal medical imaging fusion enhances diagnostic accuracy compared to unimodal approaches by providing richer contextual information about tumor morphology and biology Abdullakutty et al. (2024); Figueiras et al. (2025); Guo et al. (2017). Deep learning-based methods have accelerated progress by enabling complex feature extraction across modalities; however, their success depends heavily on dataset quality/diversity and robust external validation Houssein et al. (2021); Jha et al. (2025). Explainable AI frameworks are crucial for clinical adoption but require careful design to avoid overwhelming clinicians or diminishing trust through poorly designed explanations Abdullakutty et al. (2024); Hou et al. (2025).

Despite promising results, current fusion approaches face several limitations. Most studies rely on small private datasets, limiting reproducibility and clinical generalizability. Fusion architectures are often evaluated on single-center cohorts, and domain shift remains a critical challenge. Moreover, interpretability of multimodal modelsremains limited, hindering clinical adoption. Finally, regulatory and ethical considerations, including data privacy and bias in multimodal datasets, are rarely addressed in existing literature.

According to the LR analysis, the most common modality is the mammography MM, MRI, and US. Figure 4. Despite these advances, several gaps remain: most studies focus on two-modality fusions rather than true multi-modal integration; public datasets are lacking for many combinations; real-world deployment is hindered by interoperability issues between hospital systems; privacy concerns limit data sharing; and there is a need for prospective multicenter trials demonstrating clinical benefit over standard care Figueiras et al. (2025); Jha et al. (2025). Multimodal medical imaging fusion represents a significant advance in breast cancer detection—offering improved diagnostic performance through complementary information integration—but widespread clinical adoption requires overcoming challenges related to data standardization/interoperability, external validation across populations/institutions, explainability/trustworthiness of AI models, privacy-preserving collaboration frameworks like federated learning/XAI integration.

Table 4 show the matrix of research distribution efforts across fusion strategies and clinical tasks. While diagnostic applications dominate, significant gaps exist in explainable AI integration and workflow implementation, highlighting the need for clinically translatable multimodal systems.

Most research focuses on improving diagnostic accuracy using traditional imaging fusion, while critical aspects such as explainability, clinical integration, and real-world deployment remain largely unexplored

# 5 Conclusion

This work, analyzed the role of multimodal medical imaging fusion in improving the detection and characterization of Breast Cancer. By integrating traditional and advance modalities with emerging approaches like Microwave Imaging, multimodal fusion has demonstrated significant potential in enhancing diagnostic accuracy and reducing uncertainty. In particular, the adoption of Deep Learning and transformerbased architectures has improved performance in tasks such as classification, segmentation, and prognosis prediction. Furthermore, the integration of imaging data with clinical and genomic information is driving the emergence of Radiogenomics, supporting more precise and personalized cancer management. The distinction between traditional and advanced breast imaging modalities is continuously evolving as technological innovations introduce new acquisition techniques, enhanced protocols, and AI-driven analytical capabilities. Consequently, modern breast cancer imaging should be viewed as a dynamic and progressively integrated field in which multimodal approaches increasingly contribute to more accurate, personalized, and clinically effective diagnosis.

Multimodal fusion represents a promising and rapidly evolving paradigm in breast cancer imaging, with the potential to transform early detection, diagnosis, and treatment planning.

**Table 3** Key claims and supporting evidence identified in LR papers.

| Claim | Evidence Strength | Reasoning | Papers |
|---|---|---|---|
| Multimodal image fusion improves diagnostic accuracy over unimodal | Strong | Multiple reviews/metaanalyses show higher sensitivity/specificity when combining modalities | Abdullakutty et al. (2024); Figueiras et al. (2025); Guo et al. (2017); Jha et al. (2025) |
| Deep learning enables effective feature extraction/fusion | Strong | CNNs/transformers outperform traditional methods but depend on dataset quality | Abdullakutty et al. (2024); Cerdas et al. (2025); Houssein et al. (2021) |
| Explainable AI increasesclinician trust/adoption | Moderate | XAI tools like GradCAM/LIME/SHAP help interpret model outputs | Abdullakutty et al. (2024); Hou et al. (2025) |
| Integration with non-image data remains underexplored | Moderate | Few studies combine genomics/clinical records with images | Abdullakutty et al. (2024) |
| Lack of standardized public datasets limits generalizability | Moderate | Most studies use private/single-institution datasets | Mashekova et al. (2025) |
| Real-world workflow integration faces technical/interoperability barriers | – | Limited evidence from prospective multicenter trials | Figueiras et al. (2025) |

Future research should focus on large-scale validation, integration of heterogeneous data sources, and translation of these technologies into real-world clinical practice. This comprehensive review of traditional and advanced breast cancer detection modalities is essential for identifying complementary imaging characteristics, selecting optimal fusion strategies, and designing robust deep learning models that are both clinically relevant and scientifically novel. The future research should therefore prioritize standardized and diverse datasets, robust multimodal fusion strategies, external and multicenter validation, explainable and privacy-preserving AI, and integration of imaging with clinical and molecular information to achieve reliable and clinically applicable breast cancer diagnostic and prognostic systems.

**Table 4** Research matrix coverage by modality combination/outcome versus literature review (LR) study attributes of consensus.

| Topic/Outcome | Mammography-US Fusion | MRI/PET Fusion | Multimodal+Nonimage Data | Public Dataset Studies |
|---|---|---|---|---|
| Diagnostic Accuracy | **12** | **7** | **2** | **4** |
| Prognosis/Treatment Planning | **4** | **5** | **2** | **-** |
| Explainable AI Integration | **2** | **-** | **3** | **-** |
| Workflow Implementation | **-** | **-** | **-** | **-** |

**Supplementary information.** The Article accompanying supplementary file/s state here: - Table for the row information and the LR extraction information - Table for the journal the citation ranking for the LR articles used in this work - Sankey graph Fig 3 as .png file.

**Acknowledgements.** The authors would like to thank the S˜ao Paulo Research Foundation (FAPESP), UNESCO–TWAS, and the Brazilian National Council for Scientific and Technological Development (CNPq) for their financial and institutional support.

# Declarations

- Funding: This research was supported by the S˜ao Paulo Research Foundation (FAPESP), Brazil, through the CEPID (Research, Innovation and Dissemination Centers) program in collaboration with UNESCO–TWAS under the Postdoctoral Fellowship Program (grant numbers 2024/08242-8 and 2013/07375-0). Additional partial support was provided by the FAPESP/CePEX – S˜ao Paulo Data Integration Center for Monitoring Extreme Events (grant number 2025/07171-2) and by the Brazilian National Council for Scientific and Technological Development (CNPq) (grant number 308529/2021-9).
- Conflict of interest/Competing interests (check journal-specific guidelines for which heading to use) : The authors declare that they have no conflict of interest.
- Ethics approval and consent to participate : Not applicable
- Consent for publication : Not applicable
- Data availability : Not applicable
- Materials availability: Not applicable
- Code availability : Not applicable

- Author contribution : JA conceptualize, conduct the literature review and drafted the manuscript. JPP supervised the study and provided guidance throughout the research process. All authors reviewed the manuscript, contributed to its revision, and approved the final version.